\documentclass[%
 reprint,
 amsmath,amssymb,
 aps,
]{revtex4-2}

\usepackage{graphicx}% Include figure files
\usepackage{dcolumn}% Align table columns on decimal point
\usepackage{bm}% bold math
\usepackage{xcolor}

\begin{document}

\preprint{APS/123-QED}

\title{Detecting chaos in lineage-trees:\\A deep learning approach}

\author{Hagai Rappeport}
 \email{hagai.rappeport@mail.huji.ac.il}
\affiliation{Racah Institute of Physics, Hebrew University, Jerusalem, Israel}
\affiliation{The Rachel and Selim Benin School of Computer Science and Engineering, Hebrew University, Jerusalem, Israel}

\author{Irit Levin Reisman}
\affiliation{Racah Institute of Physics, Hebrew University, Jerusalem, Israel}

\author{Naftali Tishby}
\affiliation{The Rachel and Selim Benin School of Computer Science and Engineering, Hebrew University, Jerusalem, Israel}

\author{Nathalie Q. Balaban}
 \email{nathalie.balaban@mail.huji.ac.il}
\affiliation{Racah Institute of Physics, Hebrew University, Jerusalem, Israel }
\date{\today}%

\begin{abstract}
Many complex phenomena, from weather systems to heartbeat rhythm patterns, are effectively modeled as low-dimensional dynamical systems. Such systems may behave chaotically under certain conditions, and so the ability to detect chaos based on empirical measurement is an important step in characterizing and predicting these processes. Classifying a system as chaotic usually requires estimating its \textit{largest Lyapunov exponent}, which quantifies the average rate of convergence or divergence of initially close trajectories in state space, and for which a positive value is generally accepted as an operational definition of chaos. Estimating the largest Lyapunov exponent from observations of a process is especially challenging in systems affected by dynamical noise, which is the case for many models of real-world processes, in particular models of biological systems. We describe a novel method for estimating the largest Lyapunov exponent from data, based on training Deep Learning models on synthetically generated trajectories, and demonstrate that this method yields accurate and noise-robust predictions given relatively short inputs and across a range of different dynamical systems. Our method is unique in that it can analyze tree-shaped data, a ubiquitous topology in biological settings, and specifically in dynamics over lineages of cells or organisms. We also characterize the types of input information extracted by our models for their predictions, allowing for a deeper understanding into the different ways by which chaos can be analyzed in different topologies.

\end{abstract}

\maketitle

%\tableofcontents

\section{\label{sec:intro}Introduction \protect}
\begin{figure*}
\includegraphics{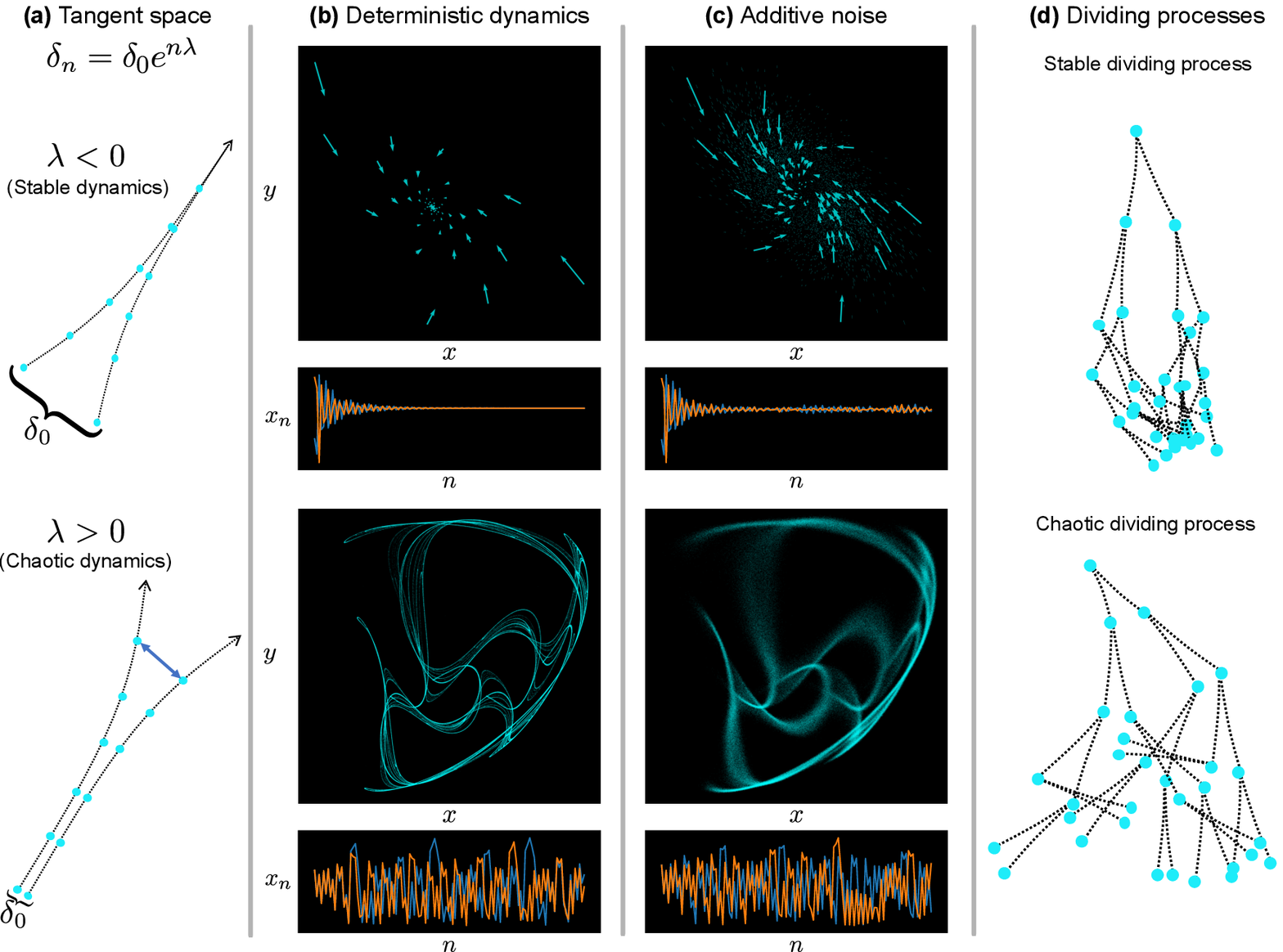}
\caption{\label{fig:intro} 
Illustrating the effects of dynamical noise on stable and chaotic dynamics.\\
\textbf{(a)} The \textit{largest Lyapunov exponent} (LLE) of a system measures the average exponential rate $\lambda$ at which trajectories which start infinitesimally close in state space ($\delta_{0}$) initially either diverge or converge, and its sign differentiates between stable (\textbf{top}) and chaotic (\textbf{bottom}) dynamics.
\textbf{(b, c)} Trajectories obtained from two quadratic maps (Supp. Eq.~\ref{eq:quad}) without \textbf{(b)} and with \textbf{(c)} additive dynamical noise where top panels depict the dynamics in 2D state space while each of the bottom panels is a 1-dimensional projection of 2 trajectories with different starting points. These systems can have very different dynamics, from convergence of all initial conditions to a fixed point in state space (top row) to the chaotic divergence of arbitrarily close initial points (bottom row), and these properties are maintained in the presence of moderate dynamical noise. \textbf{(d)} Schematic illustrations of trajectories from a system with dynamic noise (as in \textbf{(c)}) but this time for \textit{replicating} systems. At each replication, the trajectory splits into two new trajectories, resulting in a tree-like structure that can either converge (up to noise) or diverge. Inferring the LLE may be facilitated by these tree-like structures, and they are in particular relevant for many biological processes.
}
\end{figure*}

Chaos theory, which studies systems characterized by irregular dynamics and extreme dependence on initial conditions, has been instrumental in resolving the apparent paradox of complex behavior in systems governed by simple laws of motion. Many complex real-world phenomena are effectively modeled as low-dimensional dynamical systems, and so these constructs are useful in their own right, serving as more than mere toy-models. The field of Biology in particular is rich with examples of processes affected by many factors interacting in nonlinear ways, whose behavior can nevertheless be reduced to a low-dimensional dynamical description (e.g.~\cite{tyson2001network}). While all living systems depend to some degree on maintaining steady state dynamics, they are often found hazardously close to chaotic regimes~\cite{ferrell2015Embryonic}, and evidence of full-fledged chaotic behaviour or its potential have been convincingly shown in many instances~\cite{skinner1994low}, from low-dimensional systems with seemingly simple dynamics~\cite{Mosheiff2018}, through spatiotemporal chaos in multi-cellular growth programs (e.g. embryonic development~\cite{wingreen2011Embryonic}), to chaos in dynamics of cellular networks~\cite{albert2013network_reduction, glass2014dynamics}, an emerging front in modelling complex biological systems. Some of these cases may result from disregulation of control parameters, potentially leading to so called "dynamical disease"~\cite{glass1988clocks}, in which case methods from the discipline of chaos control may be used to reestablish steady state dynamics~\cite{weiss1994chaos}. Conversely, and more speculatively, transition into chaotic dynamics may be deliberate, as the increase in variability associated with such dynamics may confer benefits in certain environments. Either way, developing accurate and robust tools for detecting chaos in biological systems is of interest. While no single definition of chaos is universally accepted, a common measure of sensitivity to initial conditions in nonlinear systems is the \textit{largest Lyapunov exponent} (hereinafter LLE), which is the average exponential rate at which arbitrarily small perturbations of a system's state either increase or decrease over time (“average” over asymptotically long realizations)~\cite{strogatz2018nonlinear}. A negative value for the LLE thus implies stable dynamics and overall reduction in uncertainty while a positive value is indicative of unstable behavior (Fig.~\ref{fig:intro}(a)), and is widely used as an operational definition of chaos. When the system's equations of motion are known, a simple numerical procedure allows one to calculate its LLE~\cite{skokos2010lyapunov}. However, it is often the case when analyzing a process that one has no knowledge of the exact form of the underlying equations, and so having means of estimating a system's LLE based only on time-series it generated is important for characterizing and understanding said system.

LLE estimation from time-series in a deterministic, noiseless setting is not without its challenges, but all in all it would seem that existing methods provide satisfactory results in most cases. Alas, there are many situations in which noiselessness is not a reasonable assumption. Biological phenomena in particular, even when amenable to modeling by low dimensional dynamical systems, will almost always still contain non-negligible amounts of noise. For example, there are various ways of modeling gene expression networks as dynamical systems, but a large portion of the variability in gene expression has been shown to be intrinsic, owing to the stochastic nature of molecular interactions~\cite{Elowitz2002} and their effects on expression, such as variability in cellular growth~\cite{huh2011non}.

While \textit{measurement noise}, which obscures observation of a dynamical system's true state, will always pose a challenge for analysis, the type of noise discussed in the previous paragraph and on which we focus in this work is rather \textit{dynamical noise}. Such noise is used to model all degrees of freedom which affect a process's dynamics, but are not explicitly modeled, and are thus summarized as random variables incorporated into the equations of motion. When dynamical noise is involved, the relevant mathematical framework becomes \textit{stochastic dynamical systems} theory~\cite{arnold1974stochastic}, in which the \textit{stochastic LLE}, like its deterministic counterpart, characterizes the average rate of growth/reduction of uncertainty in the system~\cite{busse2005classification, arnold1988lyapunov}(Fig.~\ref{fig:intro}). Many biological processes can be modeled as stochastic dynamical systems, but of particular interest here are those in which the dynamics describe the change over generations of some heritable trait (the state), and sibling state variability is modeled as dynamical noise (Fig.~\ref{fig:intro}(d)). In such models, applying the equations repeatedly starting from some initial value results not in a \textit{sequence} of states, but rather in a \textit{tree}. This is a general property of replicating dynamical systems, and has many implications for analyzing the dynamics of such processes. In particular, we will later show that this unique topology harbors much information regarding a system's dependence on initial conditions.

\subsection{\label{sec:kcc}Case study - The Kicked Cell-Cycle}
To study chaotic models of biological systems, we focused in this work on the \textit{Kicked Cell Cycle} (KCC)~\cite{Mosheiff2018}, a phenomenological model proposed to explain changes in cell-cycle duration of successive generations in cancer cell lineages~\cite{sandler2015lineage}. This process is modeled as a discrete nonlinear system with additive dynamical noise, motivated by evidence that cell-cycle duration has heritable elements, and that these dynamics may be described by very few degrees of freedom. The non linearity in the equations results from coupling to an external oscillator, which models the dependence of cell-cycle duration on the phase of the circadian clock in which the cell divides (a dependence observed in many different cell types, from bacteria~\cite{Yang2010} to mammalian cells~\cite{nagoshi2004circadian}). The KCC equations are
\begin{equation}
\label{eq:KCC1}
t_{n+1}^{\pm}=t_{n}+\tau_{0}\left(1-\alpha\right)+\alpha T_{n-1}+k\sin\left(\frac{2\pi t_{n}}{T_{\text{osc}}}\right)+\xi_{n}^{\pm}
\end{equation}
\begin{equation}
\label{eq:KCC2}
T_{n}^{\pm}=t_{n+1}^{\pm}-t_{n}
\end{equation}
The model assumes a linear dependence of the cell-cycle duration at generation $n$, denoted $T_{n}$, on both the duration of the previous generation value $T_{n-1}$ and on some intrinsic duration which is a function of the cell type and environment, summarized in a single parameter $\tau_{0}$. The tradeoff between these two influences is determined by a second parameter $\alpha$. The circadian clock is modeled as an external forcing oscillator (i.e. $\sin\left(\frac{2\pi t_{n}}{T_{\text{osc}}}\right)$, where $T_{\text{osc}}$ is the  period of a full circadian cycle) and the magnitude of the coupling to this oscillator is determined by a third parameter $k$. Finally, each division results in two daughter cells which, as a result of many details of the division process, are expected to differ in their respective cell-cycle durations $T_{n}^{+}$ and $T_{n}^{-}$. This fact is modeled by additive noise, in the form of a pair of independent random variables $\xi^{+}$ and $\xi^{-}$ drawn from $\mathcal{N}\left(0,\sigma^{2}\right)$ and results, as discussed above, in the system generating tree-shaped trajectories (lineages).

Interestingly, several features of the KCCs non-linear dynamics have been observed experimentally in populations of dividing cells. In particular, the model correctly predicts multimodal division-time distributions with large variability, as well as specific non-trivial correlation patterns within lineages~\cite{Mosheiff2018, powell1955some, froese1964distribution, sandler2015lineage, martins2018cell}. As is common in many nonlinear systems, different choices for the parameters of the KCC system may give rise to very different types of dynamical behaviors. Fixed points, periodic, quasi-periodic and chaotic dynamics can all be found when traversing the system's parameter space, and these dynamical properties are largely maintained when low or even moderate noise is added to the equations (Fig.~\ref{fig:seq_kcc_results}(a)). If cells do indeed follow KCC dynamics, one immediate question is whether they likewise display different qualitative dynamical behavior.  Organisms and cells can utilize noise to their advantage in environments where increased variability may be beneficial~\cite{Rutherford1998}, and in particular cell-cycle variability has been shown to be advantageous in many scenarios (such as antibiotic exposure~\cite{balaban2004bacterial, dhar2007microbial}), but to the best of our knowledge it has never been shown that organisms can transit into chaotic dynamics as a means to the same end. The first step in investigating any such claim would have to be the development of better tools for the detection of chaos in biological systems.

\subsection{\label{sec:previous_approaches}Previous approaches}
Over the years, several methods for LLE estimation have been proposed, and we present here a brief summary before describing our approach. Some of these methods are based on the definition of the LLE as the exponential convergence/divergence rate of initially nearby trajectories (Fig.~\ref{fig:intro}(a)). Given a time-series as input, and after embedding it via time-delay into a higher dimensional space, two points which are close in space, but not in time, are treated as two different "initial conditions". The two sub-trajectories starting from these two points are compared, and the rate in which they converge or diverge is recorded. Repeating this several times and averaging the logarithms of these rates then gives an estimate for the system's LLE~\cite{wolf1985determining}. One popular representative of this class of methods, which we use throughout this work to compare our algorithm's performance to, is the one proposed by Rosenstein et al.~\cite{Rosenstein1993}, hereinafter referred to as \textit{Rosenstein's algorithm}.
Numerically calculating the LLE of a system (when the equations of motion are known) requires only access to the system's Jacobian matrix, and so another class of estimation methods attempt to fit a function to the observed dynamics and then use the Jacobian of the fitted function as an approximation of the true Jacobian. Choices for the fitted function include local linear functions~\cite{sano1985measurement}, polynomials~\cite{bryant1990lyapunov}, radial basis functions~\cite{parlitz1992identification}, neural nets~\cite{nychka1992finding} or reservoir computers~\cite{pathak2017using}. The \textit{Zero-One test for chaos}~\cite{Gottwald2004}, based on concepts from ergodic theory, does not output an LLE estimate but a binary decision of stable/chaotic for any given input time-series. Distinguishing between stable and chaotic dynamics (negative or positive LLE, respectively) is the chief motivation for estimating the LLE, and in much of what follows this distinction is our main interest. We therefore chose to use the Zero-One test for comparison purposes as well. Finally, a number of researchers have suggested deep-learning based approaches to LLE estimation~\cite{makarenko2018deep, lee2020deep, boulle2020classification}. However, in contrast to the work we describe here, all of these methods are not intended for use with data derived from noisy sources, nor data with different structure than sequential time series (or for that matter time series of different length than the ones used for training). These works are thus more suited for the theoretical study of properties of deep learning models than as candidates for detecting chaos in real-world biological data. 

\subsection{\label{sec:our_approach}The Deep Learning approach}
In recent years, the field of Machine Intelligence has experienced major breakthroughs in many disparate areas, ranging from computer vision to natural language processing, and problems on which little progress has been made over decades of intensive research are now considered solved. Much of this progress is due to advances in the study of a class of learning algorithms known as \textit{Deep Learning}. We utilized these algorithms to the LLE estimation task by casting it as a regression problem, fitting a deep learning model to map input trajectories directly to the LLE of the generating system, using computer generated trajectories and their associated (numerically calculated) LLE as input/output training pairs. To assess the resulting model's performance, it was then tested on inputs from different systems than those used for training.
For time series inputs, we used a \textit{Gated Recurrent Unit} (GRU) model (See Fig.~\ref{fig:tree_intro}(b) for a schematic illustration). Recurrent units are well suited for temporal inputs, processing them sequentially, and incrementally updating an internal representation (a \textit{hidden state}) from which the desired output is eventually computed. A major feature of deep learning techniques is their flexibility, which enables practitioners to adapt them to different input modalities and to incorporate in them various domain specific symmetries. In our case, \textit{tree-trajectories} were of special interest, and in section \ref{sec:results_3} we detail our construction of a model designed to operate on these.
For a more through technical specification of the models used and training details, refer to appendix \ref{app:dl_model}. Further analysis regarding the nature of the learned models is deferred to section \ref{sec:UC}.

\section{\label{sec:results_1}A deep-learning model can learn the relation between time-series and the largest Lyapunov exponent of the generating system \protect}

\begin{figure*}
\includegraphics{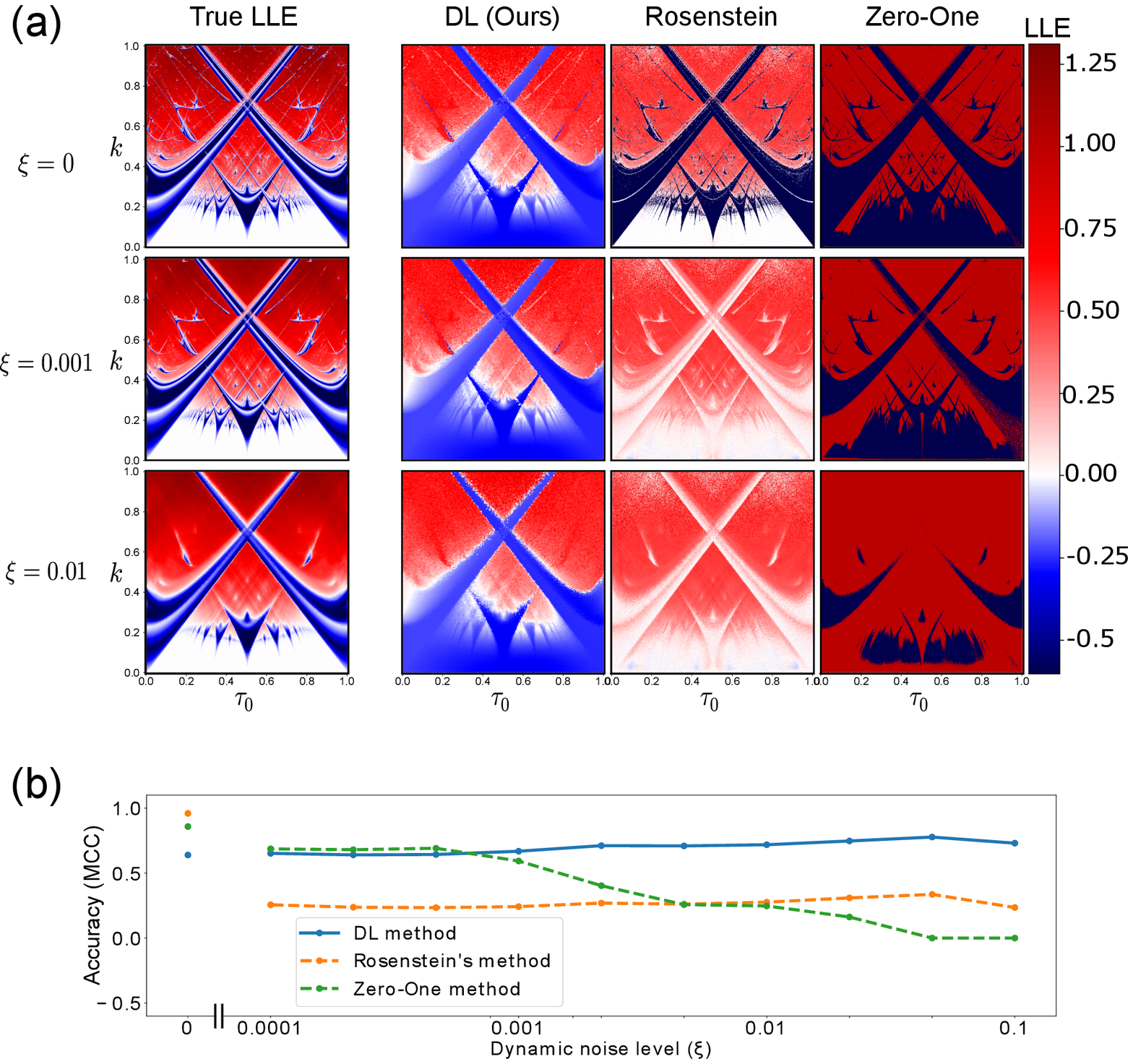}
\caption{\label{fig:seq_kcc_results} 
Comparing performance our (DL), Rosenstein's and the Zero-One methods in the task of predicting the largest Lyapunov exponent (LLE) from time series in the presence of dynamical noise.\\
\textbf{(a)} The LLE of each point in a 2D slice of the KCC parameter space (first column) with increasing levels (top to bottom) of additive noise (i.e. if the deterministic system is defined by $x_{n+1}=f\left(x_{n}\right)$ adding noise $\xi$ implies $x_{n+1}=f\left(x_{n}\right)+\mathcal{N}\left(0,\xi I\right)$). Parameters vary along axes for each of the twelve images. Blue/Red indicate negative/positive LLE values and consequentially stable/chaotic dynamics, respectively, and a LLE of 0 indicates quasi-periodic behavior. These may be compared to the corresponding prediction maps (algorithm output when the inputs are time-series derived from the respective points in parameter space). Shown here for our (DL), Rosenstein's and the Zero-One algorithms (The Zero-One algorithm outputs binary classifications)\\
 \textbf{(b)}  Performance of our (DL), Rosenstein's and the Zero-One algorithms on 11 test sets with increasing levels of additive dynamical noise each composed of 20K time-series generated from the KCC equations. Performance was quantified with the Matthews Correlation Coefficient (MCC), which measures the correlation between the true and predicted labels across a test set (1 implies a perfect classifier while 0 is equivalent to a coin toss. -1 would imply a perfectly wrong classifier)
}
\end{figure*}

Before considering tree-trajectories, we first tested the feasibility of the deep learning approach with sequential inputs, for which we could compare our results to standard LLE estimation algorithms. We trained a GRU model on KCC generated time-series of length 250 with a fixed noise level (see methods section for details regarding models and data sets). After training, we composed 11 increasingly noisy test sets by sampling 2,000 KCC parameter combinations from regions of parameter space distinct from the ones used during training. From each parameter combination, a single time-series was generated (the input for the algorithm), and the true LLE, denoted here $\lambda$, was calculated numerically. We fed these 2,000 inputs to our trained model to obtain predictions ($\hat{\lambda}$) which could then be compared to the true values. To benchmark performance, we fed the same inputs to two other widely used LLE-estimation algorithms described above - Rosenstein's algorithm and the Zero-One test. We recorded the predictions ($\hat{\lambda}$) of each algorithm and compared these to the true LLE values. While the LLE is a continuous value, it is of special interest to classify its sign, which determines whether the dynamics are stable or chaotic. To quantify models' performance as stable/chaotic classifiers, we used the Matthews Correlation Coefficient (MCC), which measures the correlation between the true and predicted labels (1 implies a perfect classifier while 0 is a equivalent to a coin toss). This analysis reveals (Fig.~\ref{fig:seq_kcc_results}) that while for purely deterministic inputs our method is slightly inferior as a classifier, its performance is stable under increasing levels of dynamical noise, and for relatively mild levels of noise ($<0.1\%$) is superior in classification ability. The difference in performance is visualized in (Fig.~\ref{fig:seq_kcc_results}(a)) by plotting the true and predicted LLE of each point in 2D parameter spaces for increasing noise levels. This visualization reveals that the Zero-One and Rosenstein's algorithms gradually lose their ability to detect stable behavior when dynamical noise is increased. The main sources of error for the Deep-Learning algorithm are the white regions in the bottom of parameter space, corresponding to quasi-periodic dynamics, which the model classifies as stable (negative LLE). Although  these dynamics have a LLE equal to 0, they are not hyper sensitive to initial conditions and so arguably should indeed be classified as stable.

\section{\label{sec:results_2} The deep-learning approach generalizes across different types of dynamical systems \protect}

\begin{figure}
\includegraphics{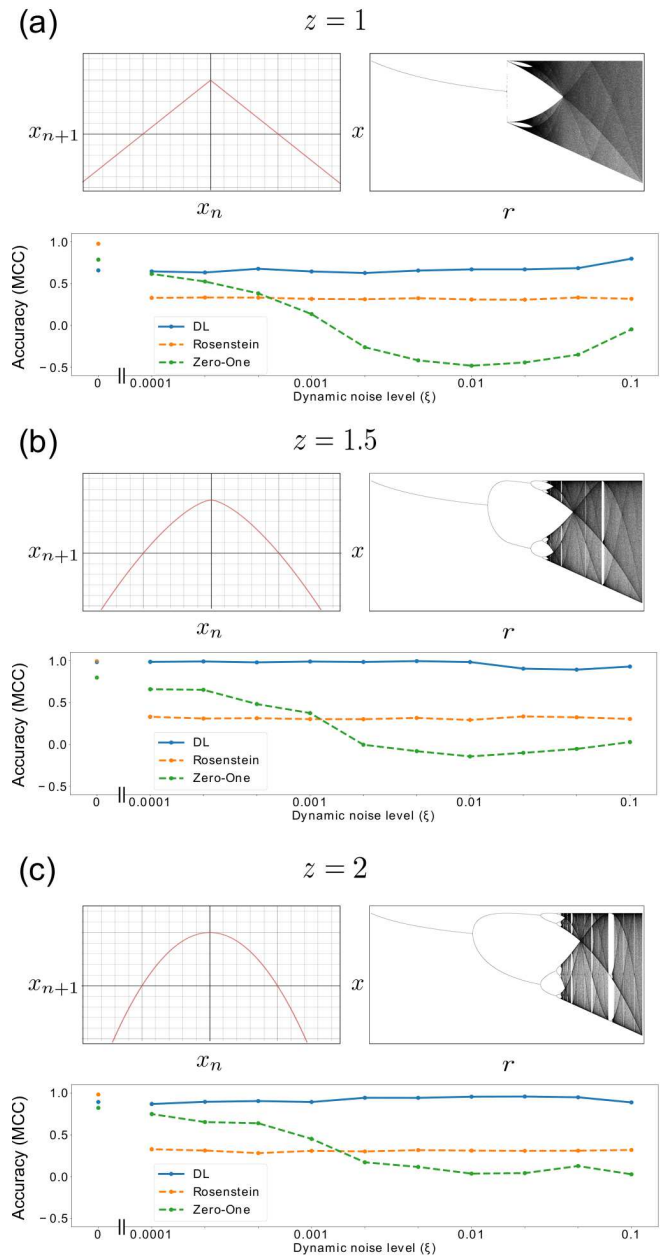}
\caption{\label{fig:seq_zmax_results} 
The deep learning approach generalizes to systems which are very different from the ones seen at train time. Here shown results on data generated from three members of the parametrized family of iterated maps $x_{n+1}=1-r\cdot\left|x_{n}\right|^{z}$ with $r\in\left[0,2\right]$ for three different choices of  $z$, which uniquely determines the systems universality class.\\
\textbf{(a)} For $z=1$, the map is similar to a tent map (\textbf{top left}), and has a similar bifurcation diagram (\textbf{top right}). The lower panel displays the performance of the Deep-Learning model (DL) trained on time-series generated from a 2-dimensional system (The KCC model) when tested on time-series generated from this 1-dimensional system. Performance in the binary classification task is evaluated by the Mathews Correlation Coefficient (MCC) and compared to the Rosenstein and Zero-One algorithms for increasing amounts of dynamical noise.  \textbf{(b, c)} Same as \textbf{(a)} but with $z=1.5$ and $z=2$ (quadratic maximum) respectively.
}
\end{figure}

Dynamical systems differ in many important aspects other than stability, and it is possible that a given LLE estimation method provides excellent results for some types of systems while failing on others, raising the need for testing on a diverse set of dynamics. This last point is especially important in a Machine Learning context, where algorithms are notorious in their tendency to \textit{overfit}. That is, to learn mappings based on properties which correlate to the desired output only in the train set, in which case there is little reason to expect generalization to new samples different then the ones seen during training. We thus added to our test sets systems which represent a wide array of dynamical systems types. It is especially illuminating in this context to observe the robustness of prediction ability with respect to differing dynamical \textit{universality classes}.

To test performance on universality classes distinct than the one used at training, we used the parameterized family of systems given by the following recurrence relation \begin{equation} \label{eq:z_map} x_{n+1}=1-r\cdot\left|x_{n}\right|^{z} \end{equation} With $r\in\left[0,2\right]$ and $x_{0}\in\left[-1,1\right]$. One can show~\cite{van1987period} that the universality class of such a system is uniquely determined by the parameter $z$, and these different universality classes correspond to qualitatively different dynamics. Prediction results on three systems from this family can be seen in Fig.~\ref{fig:seq_zmax_results}. We note that while for $z=1$ our method performs comparably to on the KCC system, for $z=1.5$ and the quadratic case $z=2$, it outperforms the other methods even in the deterministic case, and retains near perfect classification ability even with relatively high degrees of dynamical noise.

\section{\label{sec:results_3}Detecting chaos in tree-shaped data \protect}
\begin{figure}
\includegraphics{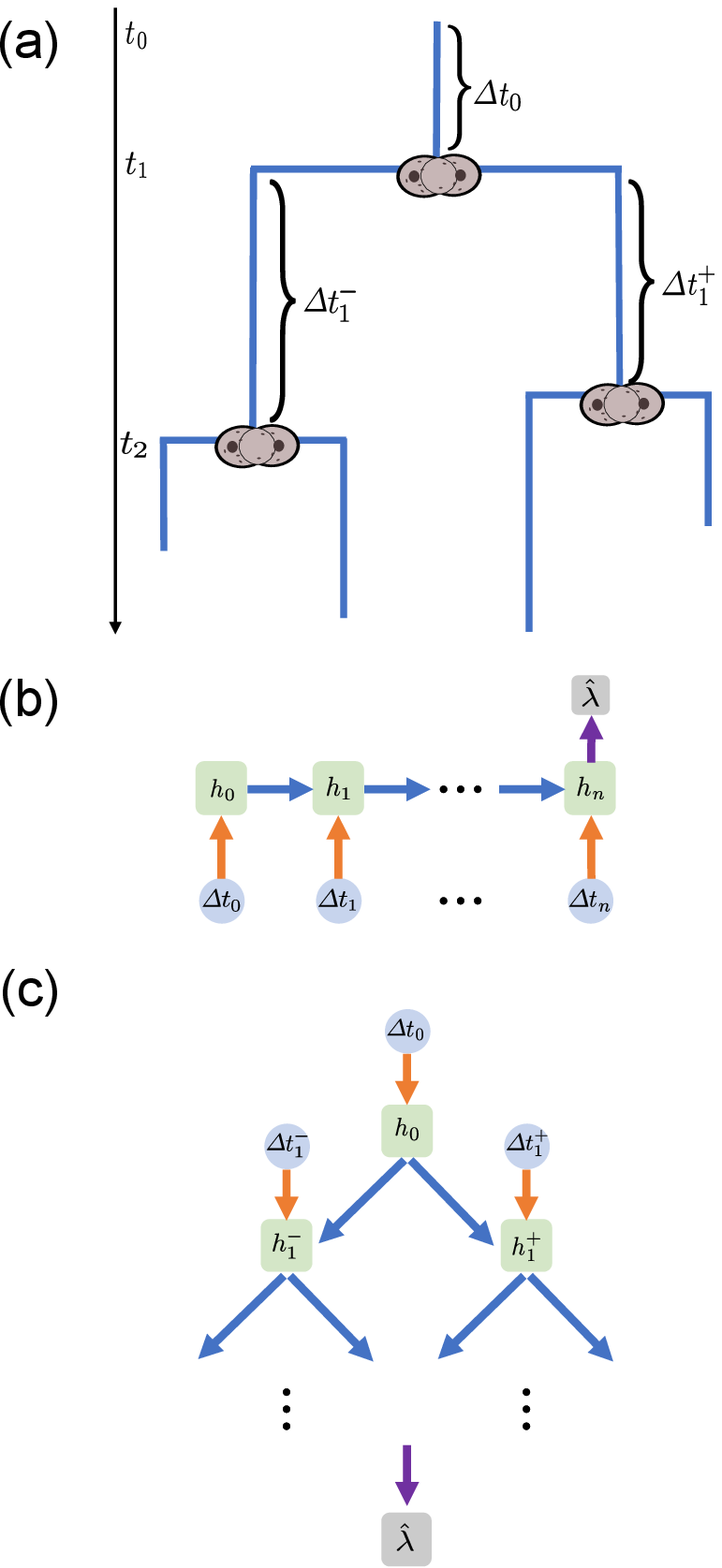}
\caption{\label{fig:tree_intro} 
Tree-trajectories generated by branching dynamical processes offer an additional dimension of information regarding the process's stability, which can be exploited by the Deep Learning approach.\\
\textbf{(a)} A schematic view of the division of cells along a lineage tree. Each cell (node in the tree) replicates after time $\Delta t_{n}$, and generates two daughter cells. Cell-cycle duration is merely one example of a heritable trait whose change over generations can be modeled as a discrete dynamical system, where variability between sibling states is modeled as instantiations of uncorrelated additive noise. Such dividing dynamical processes generate \textit{tree-trajectories} instead of standard sequential ones.
\textbf{(b)} The Deep-Learning model used hitherto for sequential data is a variant of a recurrent neural network, in which a hidden state is updated at each time step, incorporating the input at that step and the previous hidden state, and is finally mapped into the predicted output. The three mappings (colored arrows in \textbf{(b)}) are learned in training time.
\textbf{(c)} The recurrent framework can be extended to tree-trajectories, learning input-to-hidden and a hidden-to-hidden mappings and a final mapping which combines all leaf hidden-states to a prediction $\hat{\lambda}$. 
}
\end{figure}

As one would expect, all LLE estimation methods deteriorate monotonically with shorter time-series as inputs, and there is a lower limit, beneath which there is simply insufficient information in time-series to distinguish between chaotic and stable dynamics (Fig.~\ref{fig:len_vs_noise}). While some experiments can produce almost arbitrarily long sequences of measurements and for which this is not an issue, others are strictly limited in the number of data points which can be obtained. Many biological systems fall into this latter category, and in particular, tracking the fates of successive cell generations in a single lineage can become exponentially difficult for longer lineages. The realistic maximal number of generations which can be tracked is usually far less then the requirements for a reasonably reliable LLE estimation. 

The limits on recorded trajectory length in heritable cell trait dynamics may indeed pose analysis challenges, but sequential trajectories are not the natural way to view these dynamics in the first place. Namely, a cell does not simply morph into a daughter cell, but instead divides into two, and if we assume that the dynamical system accounts for the heritable portion of the trait while the two daughter cells differ merely by a noise term (as in the case of the KCC system - Eq.~\ref{eq:KCC1}), then we get tree-shaped trajectories with perturbation induced splits at every node (Fig.~\ref{fig:intro}(d)). These tree-trajectories are especially interesting in the context of measuring rates of convergence or divergence in state space, since the initial point in a tree-trajectory (the root) splits into two subtree-trajectories whose respective roots are slightly perturbed from each other (to a degree dictated by the noise's size), and these in turn each split again recursively etc (Fig.~\ref{fig:tree_intro}(a)). While a single trajectory of length $d$ may not harbor sufficient information in detecting chaotic behavior, a complete binary tree of depth $d$ has no less than $2^{2(d-2)}$ distinct pairs of trajectories of length $d-1$ whose initial points differ only by a small perturbation - which may be quite informative.

As discussed in section~\ref{sec:our_approach}, deep learning models are extremely versatile in their possible architectures. Indeed, any mapping composed of parameterized differentiable operations can in principle be trained with gradient based methods provided access to input-output samples (a train set) and a differentiable loss metric. The model we designed to analyze chaos in tree shaped data is an analog of the GRU sequence model, but while a regular recurrent model has a hidden state $h_n$ updated at each time step $n$ based on the current element in the sequence $x_{n}$ and the previous hidden value $h_{n-1}$ (Fig.~\ref{fig:tree_intro}(b)), the tree model has a hidden state updated (with the same GRU equations) based on the current \textit{tree node} and its parent's hidden state value. The final hidden states, corresponding to leaves, are then combined via a fully connected module to yield a prediction (See Fig.~\ref{fig:tree_intro}(c) for a schematic illustration). This architecture should allow the model to measure perturbation growth rates along different paths and use these to assess its LLE.

\begin{figure*}
\includegraphics{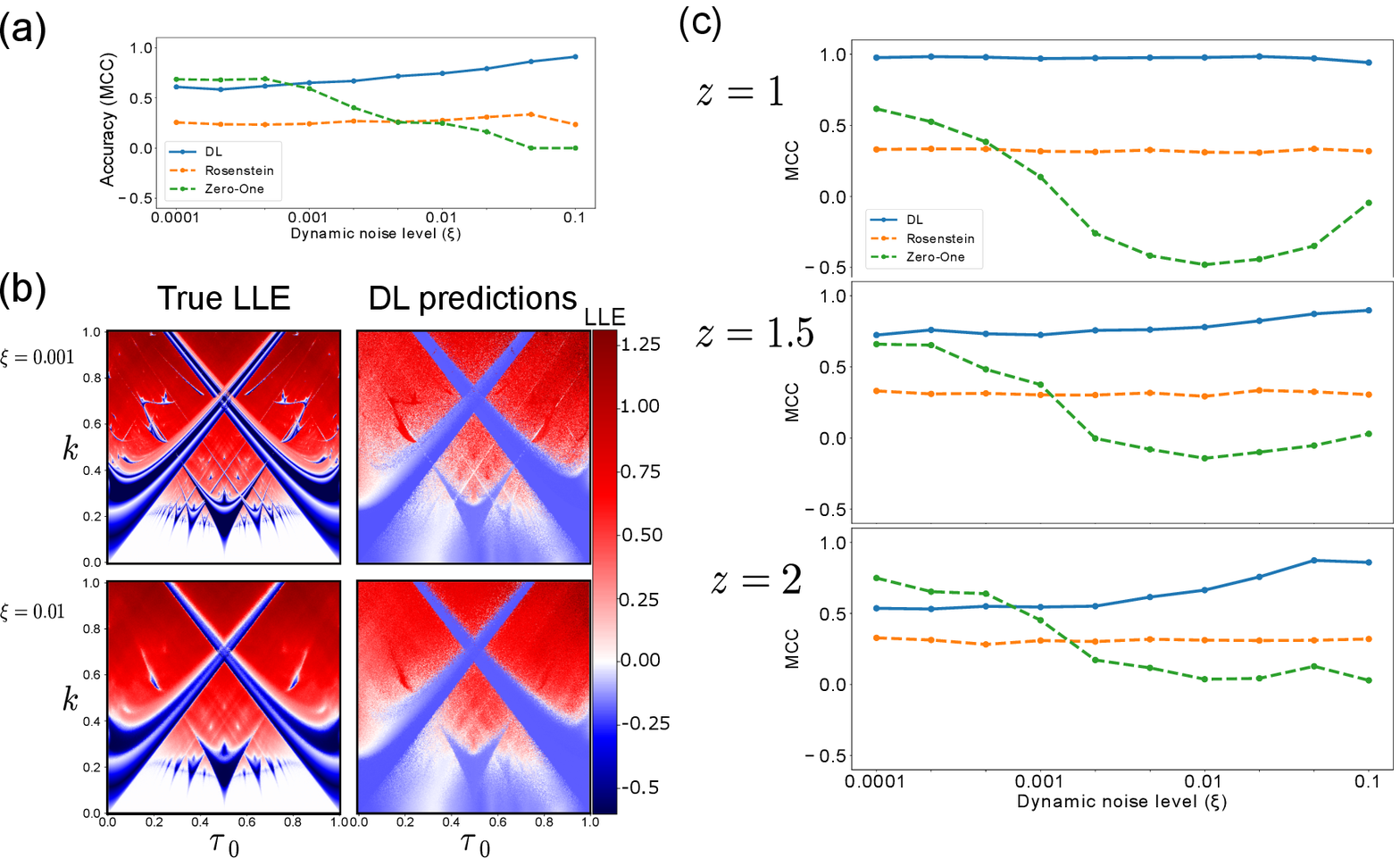}
\caption{\label{fig:tree_results} 
Detecting chaos in various dynamical systems based on tree shaped inputs.\\ \textbf{(a)} Measurements of our method's (DL) classification performance when tested on tree-trajectories of depth $8$ used as inputs, generated from a section of the KCC parameter space with increasing levels of noise (As in Fig.~\ref{fig:seq_kcc_results}). The two algorithms used for comparison were tested on sequences of comparable length (255).
 \textbf{(b)} A qualitative visualization of \textbf{(a)}, depicting predictions on sections of the KCC parameter space.
  \textbf{(c)} Performance on tree-trajectories generated from different 1-D universality classes (As in Fig.~\ref{fig:seq_zmax_results}, trajectories as in \textbf{(a)}).
}
\end{figure*}

When tested on KCC generated trees (complete binary trees of depth 8), we observed progressively increasing accuracy with higher noise levels (Fig.~\ref{fig:tree_results}(a, b)), and superior performance to Rosenstein's and the Zero-One algorithms from relatively low levels of dynamical noise. Since the latter two are not designed to take tree-trajectories as inputs, the inputs for comparison were time-series of length $255$ (the total number of time points in a binary tree of depth $8$). Note that the time-series of that length probe the dynamics over a much longer time-scale than the tree-trajectories, a significant advantage when trying to infer the LLE. Despite this advantage, the deep learning model trained on the trees-trajectories performs better in the presence of noise. 

Subjecting the tree-model to the universality class robustness test (Fig.~\ref{fig:tree_results}(c)) yielded for the quadratic map ($z=2$) similar results to the KCC generated trees, slightly better performance for $z=1.5$ and near perfect classification performance for $z=1$. Of note, as in the KCC test sets, performance increased monotonically with the amount of dynamical noise, presumably since the models are measuring perturbation growth/decrease rates (or equivalents thereof), and larger perturbations (to an extent) will make this task easier.

\section{\label{sec:UC}What have the models learned?\protect}
As discussed in section~\ref{sec:previous_approaches}, there are many different ways to infer the LLE from data, and so the question presents itself as to which of these, if any, is the one learned by our models. This question has many different aspects, but one which especially intrigued us is the inference of the LLE, a \textit{global} property of a dynamical system, from \textit{local} samples of its behavior. We were inspired by~\cite{pathak2017using}, who suggested estimating LLEs by training, for each new input, a surrogate model of the system which generated said input, and showed that these surrogates retain global properties of the system. We thus set out to characterize such global properties that our models may have learned to extract from their inputs. To this end, we chose to focus on the systems' universality class.

\begin{figure*}
\includegraphics{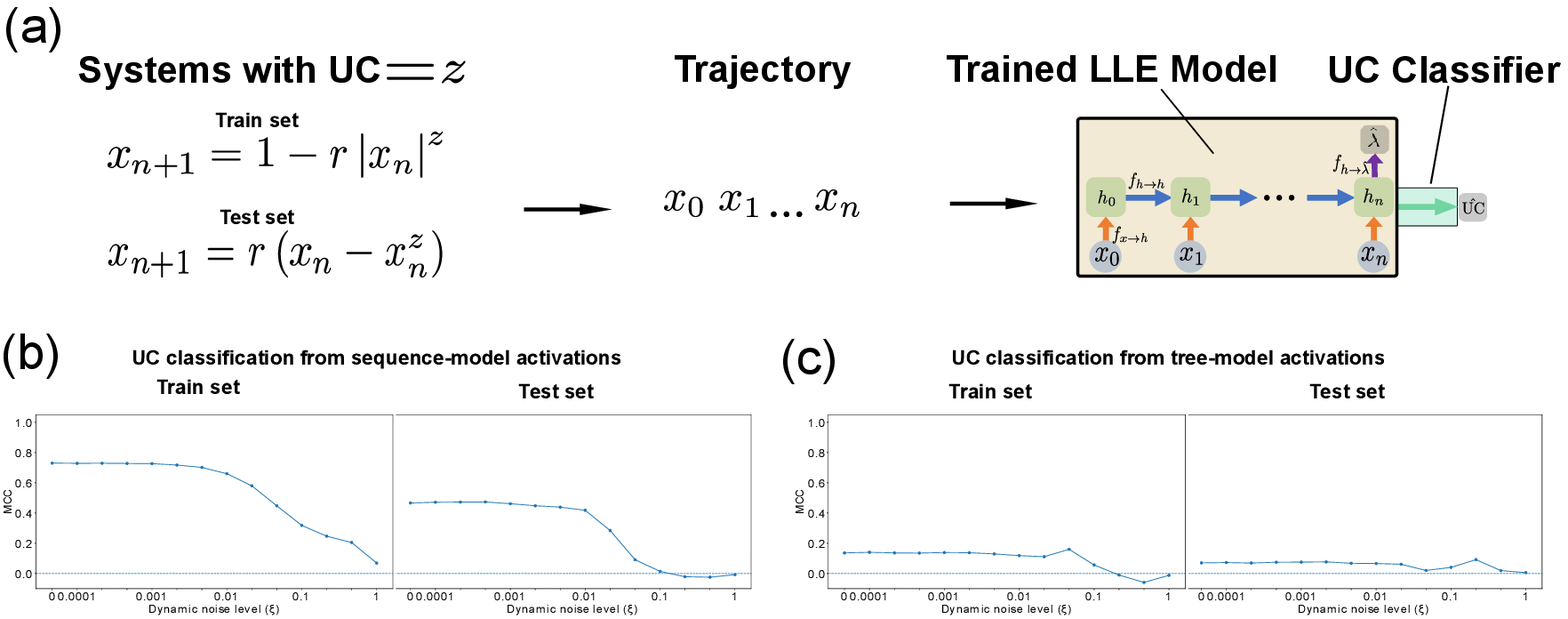}
\caption{
\label{fig:UC_predictions} 
The LLE estimation model encodes information about its input's universality class, which can be read from the model's hidden states.\\
\textbf{(a)} We fed the sequence-model, trained on KCC trajectories, with inputs generated from eq.\ref{eq:z_map} with either $z=2$ or $z=3$, a parameter choice which gives rise to two distinct universality classes. The hidden states to which the model mapped each of these trajectories ($h_{n}$) were in turn used as inputs to a simple feed-forward neural network (the UC classifier), trained to classify hidden states to the universality class of the trajectories from which they were generated.\\
\textbf{(b)} Left: UC classifier accuracy w.r.t. increasing dynamical noise on trajectories generated by eq.\ref{eq:z_map} (the same system used for training, but with different trajectory instances). Right: Same, but with trajectories generated by eq.\ref{eq:glm}, a different system which can also be parameterized to be in the same two universality classes as eq.\ref{eq:z_map}.
\\
\textbf{(c)} As in \textbf{(b)}, but for hidden-state activations from the tree-model. 
}
\end{figure*}

To investigate if and to what extent our models encode their input's universality class, we trained a new model, which takes as input a trained LLE estimation model's last hidden state $h_{n}$, obtained from a run on an input trajectory, and attempts to predict, based only on this hidden state, the universality class of the trajectory. A schematic illustration of this procedure is shown in Fig.~\ref{fig:UC_predictions}(a). More specifically, the task we chose was to differentiate between quadratic and cubic universality classes, namely discrete one-dimensional unimodal systems with maxima of leading order 2 or 3 respectively. For example, in both the system defined in eq.\ref{eq:z_map} and the one defined by  

\begin{equation}
\label{eq:glm}
x_{n+1}=r\left(x_{n}-x_{n}^{z}\right)
\end{equation}

setting the parameter z as 2 or 3 yields quadratic and cubic universality classes respectively.

We started by analyzing the \textit{sequence} model described in section~\ref{sec:our_approach} and used in the experiments mentioned hitherto, trained on KCC generated trajectories. We generated 10,000 trajectories of length 200 from eq.\ref{eq:z_map} (5,000 from each universality class) and passed them through the model to obtain 10,000 instances of the final hidden state $h_{n}$, these served as a train set. We likewise generated a test set of hidden states, obtained from eq.\ref{eq:glm}. Next, we trained a simple feed-forward neural network on the train set to classify the hidden states into the two universality classes. Results indicate that for low to moderate levels of dynamical noise, the universality class can indeed be read from the sequence-model's representation of its input (Fig.~\ref{fig:UC_predictions}(b)), suggesting that this model has learned to extract a representation of the generating dynamics. In contrast, applying the exact same technique to the \textit{tree}-model (section~\ref{sec:results_3}) we were not able to discover any encoding of the universality class in this model's final activation layer (Fig.~\ref{fig:UC_predictions}(c)), which is to be expected if this model does not build an internal representation of the dynamics, but rather measures rates of divergence/convergence more directly along different paths in the tree. 

\section{\label{sec:discussion}Discussion\protect}
Detecting the presence and degree of chaos in a system based on time-series it generated is a challenging endeavor, and is further complicated when said system is subject to random perturbations (dynamical noise). Over the decades, several approaches have been proposed to address this challenge and some have proven quite successful when tested on specific systems. But even when a given algorithm performs well on data from one dynamical system, it is highly non-trivial to extrapolate this success to other systems with different dynamical properties. We have suggested a Deep-Learning based approach to this problem, and have shown its classification abilities to be robust to dynamical noise and to varying input length. We tested the model on inputs generated from a wide array of dynamical systems, differing in their dimension, type of nonlinearity, and universality class, and observed that the performance of the learned algorithm transfers well to these different systems.

Deep learning models are highly flexible and can be constructed in different ways to operate on different types of input topologies, and so this approach lends itself naturally to data modalities beyond sequential time series. One important example is data arranged in graphs, either directed or undirected. \textit{Trees} in particular are of special interest in biological settings, as data recorded for any heritable trait in lineages will inevitably follow a tree-like structure (a binary tree in the case of simple fission, as is the case for cell lineages, or a tree of higher degree for organisms with multiple descendants). Tree shaped data seem to be especially relevant when estimating a system's sensitive dependence on initial conditions. Each time-point in a tree-trajectory (that is, a node in the tree) is followed by a split that creates two (or more) continuations of the trajectory, which may be thought of as perturbing the system in different directions and following all resulting trajectories, and then repeating this recursively for each of the newly created trajectories. Even with this intuition in mind, it is not necessarily clear exactly how much information regarding sensitive dependence is actually contained in tree-trajectories, or how one can utilize said information in this setting. By constructing Deep-Learning models designed to estimate the LLE from tree-trajectories, and showing strong classification abilities even for relatively shallow input trees (much shallower than the minimal requirement for LLE estimation from a sequential trajectory), we have provided a proof-of-concept for this idea, one that we believe warrants further research into methods for exploiting information embedded in this ubiquitous topology.

In addition to better understanding observed real-world phenomena, works such as this have at least one other important aspect - gaining insight into the mathematical models of said phenomena. Characterizing what properties of a dynamical system can, in principle, be inferred and under what conditions, enriches understanding of these properties, e.g. the non-trivial fact that the LLE can be inferred from relatively shallow tree structures. Furthermore, the use of general learning algorithms allows studying what information correlates to being able to infer any given property of the system, further advancing understanding of their interplay. The work presented in section \ref{sec:UC} offers some preliminary examples as to how such an investigation may be conducted in the case of deep learning algorithms, and while by no means comprising a complete characterization of the learned models, we believe these results offer valuable hints. 

Our ultimate objective in developing this approach was to be able to detect chaos in empirical measurements of biological processes, but both the conceptual and practical limitations that still stand in the way to this goal must be acknowledged. On the theoretical front, it is not always clear when and to what extent mathematical properties of a phenomenological model actually exist in the process being modeled. But even if we believe that a given real-world process can be chaotic for certain values of its control parameters, it is not obvious that we can extrapolate the success of our chaos detection algorithm on synthetic data and expect it to perform equally well on real data. If an algorithm such as ours outputs a positive LLE estimation for some experimental time-series, it would definitely count as \textit{evidence} for chaotic behavior, but not necessarily \textit{strong} evidence. One way to circumvent this problem is to estimate the LLE for many different values of the control parameters. If the emerging pattern of LLE values is consistent with the one expected from the model, this may count as strong evidence of chaotic behavior. These sort of experiments could also boost our confidence in the ability of the algorithm to detect chaos in any single given experimental dataset.

\subsection{\label{sec:acknowledgements}ACKNOWLEDGMENTS}
We wish to thank Zohar Ringel, Omri Abend, Shmuel Rubinstein and Mor Nitzan for helpful discussions and comments. This work was supported by the European Research Council (Consolidator grant no. 681819), the Israel Science Foundation (grant no. 597/20) and the Minerva Foundation.

\appendix

\begin{figure}
\includegraphics{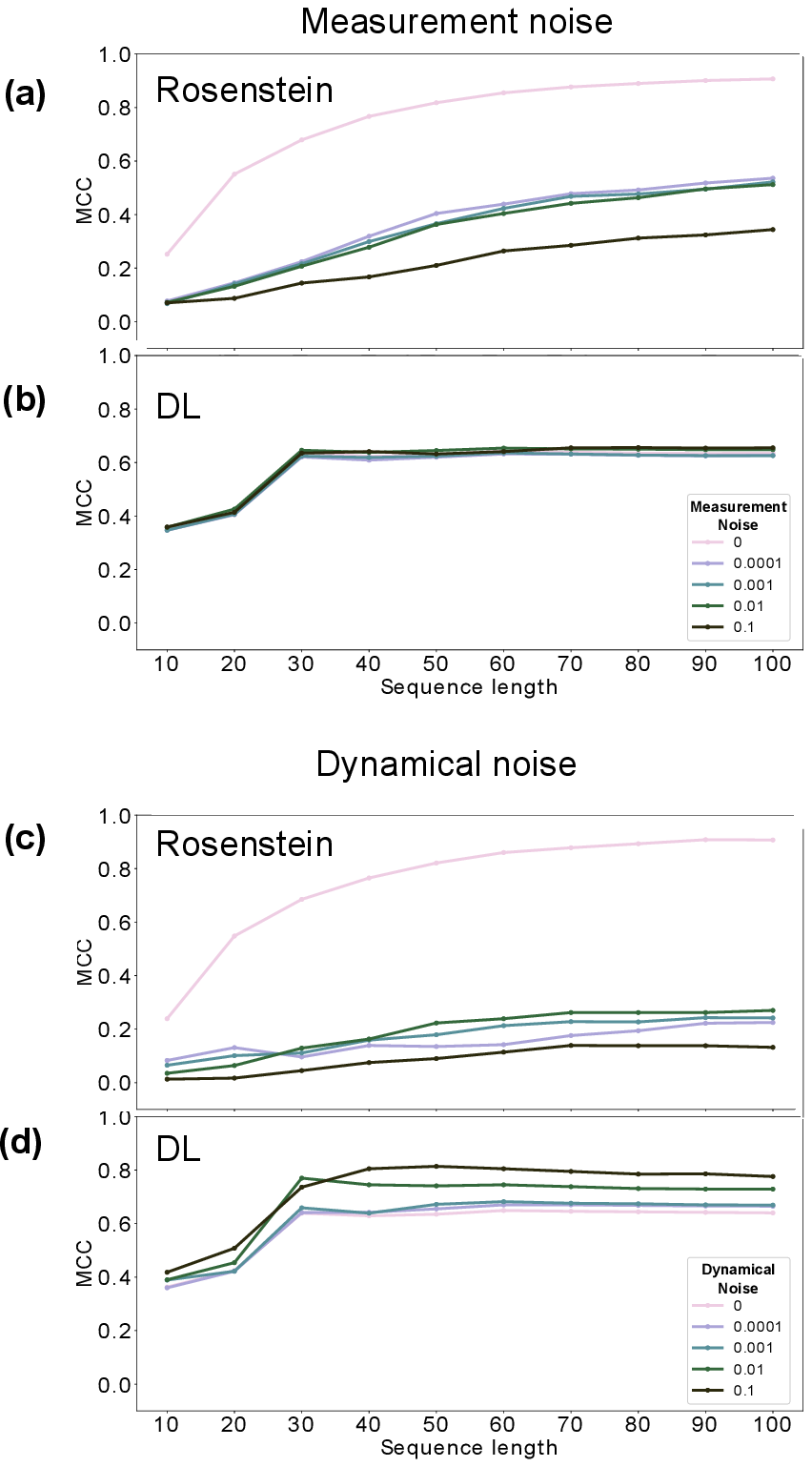}
\caption{\label{fig:len_vs_noise} 
Performance averaged over all test sets (measured by MCC, see text) of our Deep Learning model (DL) and Rosenstein's algorithm on varying input time-series length, noise type and noise level.\\
\textbf{(a, b)} Rosenstein's and the DL methods' performance when the noise in question is measurement noise (i.e. deterministic time-series are generated, and then noise in the indicated level is added).\\
\textbf{(c, d)} As in \textbf{(a, b)}, but with data generated by systems with dynamical noise, and performance measured on the task of predicting the \textit{stochastic} LLE.}
\end{figure}

\section{
\label{app:dl_model}
Deep Learning models}

\subsection{\label{sec:seq_model}Sequence model}
\subsubsection{Model}
For time-series inputs, we used a Gated Recurrent Unit (GRU) model~\cite{cho2014learning} with 64 hidden units. The model was trained using the Adam optimization algorithm~\cite{kingma2014adam} with a batch size of 32 and a learning rate of 0.0005 for 200 epochs. Regularization ($\ell_2$) was set at 0.0003 and a recurrent dropout of 0.3 was also used in training.

\subsubsection{\label{sec:seq_train_set}Train set}
 The train set was composed of 8,000 time-series of length 250 generated with the KCC equations with additive dynamical noise drawn from $\mathcal{N}\left(0,0.001\right)$. Parameter values used were $\alpha=-0.5$ while  $k$ and $\tau_0$ were sampled randomly for each trajectory, as were the initial conditions. The parameters $k$ and $\tau_0$ were sampled only from ergodic regions of parameter space (ones with a single attractor, independent of initial state). Parameter sampling was performed so as to have the distribution of train set LLEs approximately uniform between the arbitrarily chosen bounds $-2$ and $2$. 

\subsection{\label{sec:tree_model}Tree model}
\subsubsection{Model}
The model we constructed to process inputs with a (possibly incomplete) binary tree topology is schematically depicted in Fig.~\ref{fig:tree_intro}(c). 
The general approach was to have a hidden state updated at each node based on said node's value and its parent's hidden state (using GRU update equations. Specifically, a GRU unit identical to the one described in section~\ref{sec:seq_model}. This, in effect, means that for a tree with $\ell$ leaves, each of the $\ell$ paths which start from the root and end in a leaf is equivalent to a GRU run on said path, and so a run ends with $\ell$ distinct hidden states $\left\{ h_{i}\right\} _{i=1}^{\ell}$.
Next, each of these hidden states was embedded via an embedding module, a feed forward neural network (NN) $f:h_{i}\to z_{i}$ and the $z_i$'s were combined via a prediction module (a single linear layer) $g:\left\{ z_{i}\right\} _{i=1}^{\ell}\mapsto\hat{\lambda}$  
The embedding modules were two-layered feed forward NNs with layers of width 32 and RELU activation functions. The model was trained using the Adam optimization algorithm with a batch size of 32 and a learning rate of 0.001 for 100 epochs.

\subsubsection{Train set}
The train set was composed of 4,000 complete binary tree-trajectories of depth 8 generated with the KCC equations with $\alpha=-0.5$ and the other KCC two parameters - $k$ and $\tau_0$ - sampled as in  section~\ref{sec:seq_train_set}. The train set included additive dynamical noise sampled from $\mathcal{N}\left(0,0.05\right)$ for each successive generation. 

All models were trained using the Pytorch framework~\cite{paszke2017automatic}

\section{\label{sec:quad_maps}Quadratic maps}
The illustrations in Fig.~\ref{fig:intro}(b, c) were created by iterating quadratic maps for a predetermined number of iterations and recording the trajectories. A quadratic map is a 2-dimensional discrete iterated map given by the following recurrence relation
\begin{equation}
\label{eq:quad}
\begin{array}{c}
x_{n+1}=a_{1}+a_{2}x_{n}+a_{3}x_{n}^{2}+a_{4}x_{n}y_{n}+a_{5}y_{n}+a_{6}y_{n}^{2}+\xi_{x}\\
y_{n+1}=a_{7}+a_{8}x_{n}+a_{9}x_{n}^{2}+a_{10}x_{n}y_{n}+a_{11}y_{n}+a_{12}y_{n}^{2}+\xi_{y}
\end{array}
\end{equation}
Where $x, y$ are the system's variables, the random variables $\xi_{x},\xi_{y}\sim\mathcal{N}\left(0,\sigma^{2}\right)$ are additive noise and $a_{1}\ldots a_{12}$ are parameters, different values of whom yield different types of dynamics.
The parameters used for the stable attractors (top row) were $\left(a_{1},\ldots,a_{12}\right)=\left(-1.1,-1.0,0.4,-1.2,-0.7,0.0,-0.7,0.9,0.3,1.1,-0.2,0.4\right)$ and for the chaotic attractors (bottom row) $\left(a_{1},\ldots,a_{12}\right)=\left(0.2,0.8,-0.6,-0.7,-0.3,-0.2,-0.9,-0.5,0.6,-1.2,-0.3,0.8\right)$. The deterministic case (\textbf{b}) was iterated with $\sigma=0$ and the stochastic case (\textbf{c}) with $\sigma=0.005$

\bibliography{paper}% Produces the bibliography via BibTeX.

\end{document}